\pdfoutput=1
\documentclass[11pt]{article}

\usepackage[]{EMNLP2023}

\usepackage{graphicx}
\usepackage{siunitx,booktabs,caption}

\usepackage{times}
\usepackage{latexsym}

\usepackage[T1]{fontenc}
\usepackage[utf8]{inputenc}

\usepackage{microtype}

\usepackage{inconsolata}

\title{Yet Another Model for Arabic Dialect Identification}

\author{Ajinkya Kulkarni \\
  MBZUAI, UAE \\  \texttt{ajinkya.kulkarni@mbzuai.ac.ae} \\\And
  Hanan Aldarmaki \\
  MBZUAI, UAE \\  \texttt{hanan.aldarmaki@mbzuai.ac.ae} \\}

\begin{document}

\maketitle
\begin{abstract}

In this paper, we describe a spoken Arabic dialect identification (ADI) model for Arabic that consistently outperforms previously published results on two benchmark datasets: ADI-5 and ADI-17. We explore two architectural variations: ResNet and ECAPA-TDNN, coupled with two types of acoustic features: MFCCs and features exratected from the pre-trained self-supervised model UniSpeech-SAT Large, as well as a fusion of all four variants. We find that individually, ECAPA-TDNN network outperforms ResNet, and models with UniSpeech-SAT features outperform models with MFCCs by a large margin. Furthermore, a fusion of all four variants consistently outperforms individual models. Our best models outperform previously reported results on both datasets, with accuracies of 84.7\% and 96.9\% on ADI-5 and ADI-17, respectively.  
%Arabic, classified as a Semitic language, is spoken in various dialects across the Arab world. Thus, it becomes imperative to understand the inherent dialectal variations within spoken Arabic to optimize the performances of speech processing applications. In this work, we investigated ResNet and ECAPA-TDNN deep neural network architectures, and Arabic dialect identification (ADI) as a discriminative task. Furthermore, we have conducted experiments involving two distinct feature extraction methodologies, namely Mel-frequency cepstral coefficients (MFCCs) and the UniSpeech-SAT large acoustic model. Thereafter, we created a fusion system by weighted sum of scores obtained from all the proposed systems. To validate the effectiveness of our proposed approach, we have assessed the system's performance on the third and fifth editions of the Multi-Genre Broadcast Challenge (MGB-3 and MGB-5) for the ADI task. The obtained results showed better performances in comparison to systems submitted to MGB-3 and MGB-5 challenges.

\end{abstract}

\section{Introduction}

Dialect identification can be viewed as a special case of language recognition \cite{intro1,intro2}. 
Both tasks suffer from similar performance issues in the presence of background noise, channel mismatch, prosodic fluctuations, and so on. However, with closely related dialects having a small difference in both acoustic and linguistic feature space, dialect identification tasks are substantially more difficult in nature \cite{intro3}. %These factors make dialect identification a more complex task to solve. 
The Arabic language is spoken in various dialects across the Arab world, in addition to Modern Standard Arabic (MSA) which is used in official and educational settings. Speech recognition systems trained on MSA data generally don't generalize well to dialectal Arabic and specialized dialectal models may be needed for improving automatic speech recognition (ASR) performance in systems developed for specific populations. Dialect identification could facilitate the development of dialectal speech recognition systems in various ways, such as by identifying dialectal utterances in large multi-dialectal corpora, or online dialect identification for routing utterances to dialect-specific ASR modules. 

To enable the development of spoken Arabic dialect identification systems, two benchmark datasets have been developed: ADI-5, which was deployed as part of the MGB-3 challenge \cite{mgb3} and ADI-17, deployed as part of the MGB-5 challenge \cite{mgb5}. For both challenges, the top systems developed and submitted for the initial challenges remain the best performing systems reported in the research literature for these benchmarks. The ADI-5 training set consists of 10 hours of dialectal speech from broadcast news, covering five dialects: Egyptian (EGY), Levantine (LAV), Gulf (GLF), North African (NOR), and Modern Standard Arabic (MSA), in addition to two hours each for development and test sets. The ADI-17 data set consists of 17 dialectal classes for a total of 3K hours extracted automatically from YouTube. Roughly 58 hours of data were manually verified for the development and test sets.

In this paper, we describe spoken dialect identification models we developed and tested on these benchmarks, and we report results exceeding the best performing models submitted to both challenges. 
We experimented with the Residual networks (ResNet) \cite{resnet} and Emphasized Channel Attention, Propagation and Aggregation (ECAPA-TDNN) \cite{ecapatdnn} architectures. Both architectures have been successfully employed for speaker verification tasks. In addition, ResNet was used in the best performing dialect identification system in the MGB-5 challenge, and ECAPA-TDNN has been recently explored for dialect classification, as in \citet{lonergan2023towards} for Irish dialects. In addition, we explored the use of acoustic features extracted from the UniSpeech-SAT \cite{unispeechsat} model, which have been shown to provide improvements in various tasks in the SUPERB benchmark \cite{yang2021superb}. We observe large improvements in accuracy by incorporating these features into our models. We also employ data augmentation via additive noise and speed perturbation, which generally help improve the generalization of speech classification models. Our best model result is 84.7\% accuracy in the ADI-5 test set, compared to 75\% previously reported as the best result in \citet{mgb3}. In ADI-17, our best model achieves 96.9\% accuracy compared to 94.9\% previously reported as the best model in \citet{mgb5}.

%toolkit\footnote{\url{https://kaldi-asr.org}} for data augmentation. To demonstrate the robustness of our proposed work, we demonstrated the performance of the proposed systems on the third and fifth rounds of the Multi-Genre Broadcast Challenge (MGB-3, MGB-5).

%The paper is organized as follows: Section 2 describes previous work on Arabic dialect identification. The details of the proposed model are outlined in Section 3; Section 4 describes the experimental settings and the results are presented in Section 5. 

\section{Related Work}

In this section, we describe the approaches proposed for ADI tasks in MGB-3 and MGB-5 challenges, which are used as baseline systems in this work. We first describe the top two performing systems for the MGB-3 challenge (ADI-5) \cite{mgb3}, followed by the top two systems in the MGB-5 challenge (ADI-17) \cite{mgb5}. 

The MIT-QCRI ADI system \cite{mitqcri1,mitqcri2} combines acoustic and linguistic features within a Siamese neural network framework to reduce dimensionality based on i-vectors. They used loss functions involving both Euclidean and cosine distances and employed support vector machines as the backend classifier. In contrast, the University of Texas at Dallas (UTD) submission \cite{utd} to the MGB-3 challenge fused five systems, incorporating acoustic and lexical information through various techniques, including i-vectors, Generative Adversarial Networks (GANs), Gaussian Back-end (GB), and BNF i-vector features. The UTD system obtained the second-best performance with an overall accuracy of 70.38\% \cite{mgb3}.

Duke Kunshan University (DKU) submitted four variants of ResNets with different block sizes and datasets, which were fused to achieve the best performing system in the MGB-5 challenge \cite{mgb5}. The DKU system employed a ResNet with global statistics pooling and a fully connected layer. They used the Kaldi toolkit for data augmentation, including speed-perturbation and datasets such as MUSAN and RIR. The ResNet system was trained using cross-entropy loss with a softmax layer, taking 64-dimensional mel-filterbank energy features as input. On the other hand, the University of Kent (UKent) MGB-5 system \cite{ukent} used a neural network architecture combining Convolutional Neural Networks (CNN) and Long Short-Term Memory (LSTM) networks with Time-Scale Modification (TSM). The UKent system reported an accuracy of 93.1\% on the test set. 

While the best performing models reported in the original MGB-3 and MGB-5 challenges have not been outperformed in later publications (to the best of our knowledge), several other studies proposed model variants and analyzed the performance in various ways. Regarding the use of pre-trained self-supervised acoustic models, \citet{sullivan23_interspeech} recently utilized the XLS-R model \cite{babu2021xls}, which is a multi-lingual pre-trained acoustic model that includes Arabic as one of the languages used in pre-training, and HuBERT \cite{hubert}, which was pre-trained solely in English. They fine-tuned dialect classification models on the ADI-17 dataset, and interestingly, the model based on HuBERT outperformed the XLS-R-based model, in spite of the multi-lingual pre-training of the latter. This indicates that the quality of the features extracted from pre-trained acoustic models may depend more on the self-supervised training details rather than linguistic coverage. A model out-performing HuBERT on several benchmark tasks is the UniSpeech-SAT acoustic model \cite{unispeechsat}, which includes additional objectives on top of the HuBERT model to facilitate speaker-aware representations, which also generally embody non-linguistic characteristics of utterances, such as tone and emotion.  

\section{Proposed Model}

As the space of possible architectural or feature variations increases with the increasing volume of developments in the ML field, exhaustively searching all possible architectures is unfeasible. Therefore, we draw inspiration from the best performing models in related literature to reduce the search space and increase the likelihood of finding a best performing model. We selected two neural network architectures, ResNet and ECAPA-TDNN, for their potential in speech classification tasks. For feature extraction, we compare classical MFCC features with the pre-trained UniSpeech-SAT large acoustic model \cite{unispeechsat}  that has been shown to provide consistent improvements in various Speech classification benchmarks. Finally, as best models in previous works typically include a form of ensemble, we experimented with fusing all model variants to further improve performance. We describe the details of these parts in this section. %Thereafter, we fused the scores obtained from the softmax layer of all systems for predicting the dialect classes as a "Fusion" system.
%1. General overview of work, speech-feat-DNN-identification \\
%2. Subsection ResNet34, and ECAPA-TDNN \\
%3. Subsection feature extraction: MelFilterbank and UniSpeech \\
%4. Training details \\

\subsection{Feature extraction}

%MFCC features have been used popularly in automatic speech recognition and speaker verification tasks \cite{srmfcc,asrmfcc}. Consequently, 

We experimented with two types of features: classical acoustic features, namely MFCCs, and modern acoustic features extracted from a large pre-trained acoustic model, namely %UniSpeech-SAT \citet{unispeechsat}.  %Furthermore, we used the PyTorch Audio library to extract the MFCCs with frame-level instance normalization. 
%We also employ 
the Universal Speech representation learning with speaker-aware pre-training (UniSpeech-SAT) \cite{unispeechsat}. The large variant of this model demonstrated outstanding performance in various tasks in the SUPERB benchmark \cite{superb}, including linguistic and non-linguistic tasks, such as speaker diarization and emotion recognition. UniSpeech-SAT model is built on the HuBERT model \cite{hubert} with additional self-supervised objectives involving utterance-wise contrastive learning and utterance mixing augmentation. The speaker-aware pre-training enabled the model to improve the discriminating capabilities of embeddings learned under self-supervised learning. In total, the large variant of UniSpeech-SAT was trained on 94K hours of English speech data from various sources, including Audiobooks and YouTube. 
We extracted 1024-dimensional features from the pre-trained UniSpeech-SAT\footnote{\url{https://github.com/microsoft/UniSpeech}} model %for given input as a raw speech utterance. The 
and kept model parameters frozen. For MFCCs, we extract 80-dimensional features using a window length of 25 ms with a sliding window of 10 ms and frame-level instance normalization.

\subsection{Network architectures}
 We experimented with two network architectures that have been shown to work well in speech classification tasks: ResNet and ECAPA-TDNN, which we describe below. 
%On VoxCeleb datasets, speaker verification tasks have been extensively used with state-of-the-art results from ResNet and ECAPA-TDNN \cite{ecapatdnn}. 
%In this study, we investigated these two deep neural network architectures (ResNet and ECAPA-TDNN) utilizing input as the previously described features.

\subsubsection{ResNet}

We use the ResNet architecture \cite{resnet} as our first model. Our model is composed of four residual networks, each consisting of two convolutional layers in addition a skip connection. We utilize batch normalization and ReLU activation functions. Statistical pooling is implemented to map the variable length feature frames to a time-invariant representation by aggregating frame level mean and variance as statistical parameters. The output of statistical pooling is followed by two feed-forward layers. 
We employ the original ResNet34 set-up as described in the original paper \cite{resnet}, which has 34 2D-convolutional layers organized into 4 residual network blocks, with each block containing a specific number of layers [3, 4, 6, 3], and the convolutional filters for these layers are [32, 64, 128, 256] respectively. The last feed-forward layer includes the output dimension of a number of dialect classes to identify with Additive Angular Margin (AAM) softmax layer \cite{aam} with a scale of 30.0 and margin of 0.4, trained with cross-entropy loss function.

\subsubsection{ECAPA-TDNN}

The ECAPA-TDNN architecture \cite{ecapatdnn}, based on the x-vector architecture \cite{xvector}, utilizes a Squeeze-excitation (SE)-Res2Net module in each block. These modules consist of 1-dimensional convolutional layers, ReLU activation, batch normalization, and 1-dimensional Res2Net modules with impactful skip connections and SE blocks. This design allows the model to extract hierarchical and global information from the input features.
Additionally, the architecture incorporates attentive statistical pooling by calculating channel-dependent frame attention-weighted statistics (mean and variance). This process transforms variable-length hidden outputs into a time-invariant representation. The representation is further processed through feed-forward layers. Similar to the ResNet architecture, we use the AAM-softmax as the final layer and train it with the cross-entropy loss criterion. The model uses 512 channels in 1-dimensional convolutional layers, 128 dimensions for SE-Block and attention, and a scaling factor of 8 for each Res2Block. The output dimension for feed-forward layers is set to 192, and the last feed-forward layer's dimension corresponds to the number of dialect classes. 

\subsection{Inference Scheme}

%HANAN - started working frrom here, previous sections will be edited afterwards, 
In our model, we integrate a similarity measure with our learned classifiers to enhance classification performance \cite{lee2012enhanced, nguyen2013supervised, roul2017modified}. ResNet and ECAPA-TDNN are optimized for dialect identification via softmax, which we augment with a similarity-based measure based on the final embeddings produced by the network. For each dialect class, we randomly extract a cohort of 500 samples from the training set, and we calculate the average cosine similarity score between the test utterance and the cohort representing each class. After normalizing the scores, we combine them with the softmax scores by averaging them with equal weight (0.5) and selecting the class with the maximum score.

\section{Experimental setup}
\subsection{Datasets}

We evaluate the dialect identification model on two Arabic dialect identification tasks: the MGB-3 ADI-5 dataset \cite{mgb3}, and the fine-grained MGB5 ADI-17 dataset \cite{mgb5}. ADI-5 training set consists of 13,825 utterances (53.6 hours), and the test and development sets consist of 1,524 (10 hours) and 1,492 (10 hours) utterances, respectively, with each set having approximately 2 hours of data per dialect class: Egyptian (EGY), Levantine (LAV), Gulf (GLF), North African (NOR), and Modern Standard Arabic (MSA).
In ADI-17, approximately 3,000 hours of training data were labeled via distant supervision into 17 dialect classes using the origin country of the YouTube videos from which they were extracted. The testing and development sets contain $\sim$25 and $\sim$33 hours of speech, respectively, manually verified by human annotators. 

\subsection{Data Augmentation}

For data augmentation, we apply additive noise drawn from the Music, Speech, and Noise corpus (MUSAN) \cite{snyder2015musan} and the QMUL impulse response dataset \cite{stewart2010database}. We also apply speed perturbation, where the tempo is modified by factors of 0.9 and 1.1.  All noise augmentation was implemented using the Kaldi toolkit \cite{Povey_ASRU2011}.

\subsection{Training settings}

During the training phase, each model was initially trained with randomly selected 5-second segments from training utterances for the first 50 epochs. Subsequently, the duration of the training segments was reduced to 4 seconds for a total of 100 epochs to enable the model to generalize to short-duration utterances. All systems were trained using the Adam optimizer with a triangular learning scheduler policy and a batch size of 256. %For the inference phase, the epoch was selected based on each system's performance on the development set.

\section{Results}

%TODO add more recent models, if available
\begin{table}[!t]
\centering
\caption{Performance evaluation on MGB-3 ADI-5 test set (in \%) with baseline systems submitted to MGB-3 challenge. UniS denotes the UniSpeech-SAT feature extraction.}
\resizebox{\columnwidth}{!}{
\begin{tabular}{|l | c| c| c| c|} 
\hline \textbf{System} & \textbf{Features} & \textbf{Accuracy} & \textbf{Precision} & \textbf{Recall} \\
 \hline 

\multicolumn{5}{|l|}{Best systems from \cite{mgb3}}		\\		\hline										
MIT-QCRI	&   ---		&	75.0	&	75.1	&	75.5 \\
UTD			&   ---		&	70.4	&	70.8	&	71.7 \\ \hline
ResNet		&   MFCC	&	74.2	&	74.1	&	74.4 \\
ECAPA	    &   MFCC	&	75.3	&	75.1	&	75.3 \\
ResNet		&   UniS	&	80.4	&	80.4	&	80.5 \\
ECAPA		&   UniS	&	82.5	&	82.6	&	82.7 \\
Fusion		&   ---		&	\textbf{84.7}	&	\textbf{84.8}	&	\textbf{84.9} \\

\hline
\end{tabular}}
\vspace{-0.6em}
\label{tab:stat}
\end{table}

%TODO add more recent work, where applicable
\begin{table}[!t]
\centering
\caption{Performance evaluation on MGB-5 ADI-17  test set (in \%) with baseline systems submitted to MGB-5 challenge. UniS denotes the UniSpeech-SAT feature extraction.}
\resizebox{\columnwidth}{!}{
\begin{tabular}{|l | c| c| c| c|} 
\hline \textbf{System} & \textbf{Features} & \textbf{Accuracy} & \textbf{Precision} & \textbf{Recall} \\
 \hline 

\multicolumn{5}{|l|}{Best systems from \cite{mgb5}}		\\		\hline	
DKU	            &   ---		&	94.9	&	94.9	&	94.9 \\
UKent			&   ---		&	91.1	&	91.1	&	91.1 \\  
\hline
%\multicolumn{5}{|l|}{Models from \cite{sullivan23_interspeech}}		\\		\hline
%HuBERT-17			&   HuBERT		&	92.2	&	---	&	--- \\  
%xls-r-300m-17			&   XLS-R		&	90.8	&	---	&	--- \\  

%\hline
ResNet		    &   MFCC	&	90.1	&	90.1	&	90.1 \\
ECAPA	        &   MFCC	&	92.2	&	92.2	&	92.2 \\
ResNet		    &   UniS	&	95.7	&	95.7	&	95.7 \\
ECAPA		    &   UniS	&	96.1	&	96.1	&	96.2 \\
Fusion		    &   ---		&	\textbf{96.9}	&	\textbf{96.9}	&	\textbf{96.9} \\

\hline
\end{tabular}
}
\vspace{-1.1em}
\label{tab:stat1}
\end{table}

Tables \ref{tab:stat} and \ref{tab:stat1} show the performance of our model variants in ADI-5 and ADI-17 test sets, respectively. \textit{Fusion} refers to an ensemble model where scores from all four variants are combined, each with an equal weight of 0.25. We also show the performance of the best performing models from the original challenges, which have not been previously outperformed to the best of our knowledge. 

We observe consistent results in both datasets: ECAPA-TDNN network consistently outperforms ResNet, and the models using UniSpeech-SAT features consistently outperform those using MFCC features. Incorporating these pre-trained features results in 4\% to 5\% absolute improvement in accuracy for both models. We observe additional gains of 0.8\% to 2\% improvement in absolute accuracy by fusing all four model/feature combinations. The highest performance gain is observed by using UniSpeech-SAT features as input, which leads to outperforming all previous baselines. 

%The obtained results demonstrate the capability of ResNet and ECAPA-TDNN to learn the discriminative tasks on speech modality other than speaker verification.  
%Although UniSpeech-SAT-based feature extraction is primarily pre-trained in the English language, the obtained results indicate that it is useful for capturing the latent information across speech utterances for other languages concerning discriminative tasks. As in the case of MGB-5 similar channel domain exists in the training and evaluation set due to data curation from YouTube. MGB-3 has overall lower system performances compared to MGB-5 as indicated by Table 1. and Table 2.

\section{Conclusions}

This paper described variations of model architectures, namely ResNet and ECAPA-TDNN, employing two acoustic features: classical MFCCs and self-supervised UniSpeech-SAT, leading to state-of-the-art performance in two spoken Arabic dialect identification benchmarks: ADI-5, and ADI-17. UniSpeech-SAT features, which are extracted from a large pre-trained model optimized for acoustic and speaker variability, consistently demonstrated superior performance compared to MFCC features. Despite being pre-trained solely in English speech, UniSpeech-SAT illustrates transfer learning capability by extracting suitable feature representations for this discriminative task in the Arabic language. This may also indicate that non-linguistic acoustic variability (such as speaking tone, for example) could play a role in dialect identification. Consistent with previous models from the MGB-3 and MGB-4 challenge, fusing multiple models results in consistent improvements of overall performance.

\section{Limitations}
In this work, we limited our analysis and exploration to two network architectures and two types of acoustic features. We based our choice on observations from the current literature on dialect identification, speech classification, and self-supervised acoustic models. However, many additional features and architectural variations could have been explored, with additional detailed analysis of the different combinations. Furthermore, we did not analyze the acoustic features that are most discriminative in these datasets, which is a complex analysis that eludes us at this stage, but future work could explore more on which aspects of an utterance (linguistic, tonal, other) are most useful for dialect identification. 

%TODO mention other features and analysis of the repsentations learned by the model. 

%In the future, we would like to explore large acoustic models such as WaveLM, and HuBERT for Arabic dialect identification tasks. Moreover, it will be pertinent to exploit the existing unlabelled Arabic dialectal speech resources under the framework of self-supervised learning to further enhance the performances with large acoustic models. 

\bibliography{anthology,custom}
\bibliographystyle{acl_natbib}

\end{document}